\documentclass{article}
\usepackage{spconf,amsmath,graphicx}
\usepackage{numprint}
\usepackage{tabularx}
\usepackage{dsfont}
\usepackage{mathtools}
\usepackage{url}

\DeclarePairedDelimiter{\norm}{\lVert}{\rVert}

\newcommand{\system}[1]{{\small \textsc{#1}}}
\newcommand{\tablesystem}[1]{\textsc{#1}}

\usepackage{microtype}
\usepackage{booktabs}
\usepackage{tabularx}
\usepackage{multirow}
\usepackage{siunitx}
\newcommand{\mytable}{
    \centering
    \renewcommand{\arraystretch}{1.2}
    }
\newcolumntype{C}{>{\centering\arraybackslash}X}
\newcolumntype{L}{>{\raggedright\arraybackslash}X}

\title{Acoustic word embeddings for zero-resource languages using self-supervised contrastive learning and multilingual adaptation}
\name{Christiaan Jacobs$^1$\qquad Yevgen Matusevych$^2$ \qquad Herman Kamper$^1$\thanks{This work is supported in part by the National Research Foundation of South Africa (grant number: 120409), a Google Faculty Award, and financial support from the School of Data Science and Computational Thinking (SU).}
}
\address{$^1$E\&E Engineering, Stellenbosch University \& $^2$School of Informatics, University of Edinburgh\\
{\small \texttt{20111703@sun.ac.za, yevgen.matusevych@ed.ac.uk, kamperh@sun.ac.za}}}

\usepackage{cite}
\let\oldbibliography\thebibliography
 \renewcommand{\thebibliography}[1]{\oldbibliography{#1}
                                    \setlength{\itemsep}{-1pt}
                                    \vspace*{-0mm}}

\usepackage{xcolor}
\definecolor{mycolor}{HTML}{FF6600}
\definecolor{mycolor2}{HTML}{6699CC}
\definecolor{mycolor3}{HTML}{CC0000}

\usepackage[prependcaption,textsize=scriptsize]{todonotes}
\begin{document}
%
\maketitle
\begin{abstract}
Acoustic word embeddings (AWEs) are fixed-dimensional representations of variable-length speech segments. For zero-resource languages where labelled data is not available, one AWE approach is to use unsupervised autoencoder-based recurrent models. Another recent approach is to use multilingual transfer: a supervised AWE model is trained on several well-resourced languages and then applied to an unseen zero-resource language. We consider how a recent contrastive learning loss can be used in both the purely unsupervised and multilingual transfer settings. Firstly, we show that terms from an unsupervised term discovery system can be used for contrastive self-supervision, resulting in improvements over previous unsupervised monolingual AWE models. Secondly, we consider how multilingual AWE models can be adapted to a specific zero-resource language using discovered terms. We find that self-supervised contrastive adaptation outperforms adapted multilingual correspondence autoencoder and Siamese AWE models, giving the best overall results in a word discrimination task on six zero-resource languages.
%
%
\end{abstract}
%

\section{Introduction}
\label{sec:intro}

A \textit{zero-resource} language is one for which no transcribed speech resources are available for developing speech systems~\cite{jansen+etal_icassp13,dunbar+etal_interspeech19}.
Although conventional speech recognition is not possible for such languages, researchers have shown how speech search~\cite{levin+etal_icassp15,huang+etal_arxiv18,yuan+etal_interspeech18}, discovery~\cite{park+glass_taslp08,jansen+vandurme_asru11,ondel+etal_arxiv19,rasanen+blandon_arxiv20}, and segmentation and clustering~\cite{kamper+etal_asru17,seshadri+rasanen_spl19,kreuk+etal_interspeech20} applications can be developed without any labelled speech audio.
In many of these applications, a metric is required for comparing speech segments of different durations.
This is typically done using dynamic time warping (DTW).
But DTW is computationally expensive and can be difficult to incorporate directly into downstream systems (see e.g.\ the alterations required in~\cite{anastasopoulos+etal_emnlp16}).
\textit{Acoustic word embeddings}~(AWEs) have emerged as an alternative.
Instead of using alignment, speech segments are mapped to vectors in a fixed-dimensional space.
Proximity in this embedding space should indicate similarity of the original acoustic segments~\cite{levin+etal_asru13}.

Several AWE models have been proposed~\cite{bengio+heigold_interspeech14,he+etal_iclr17,audhkhasi+etal_stsp17,wang+etal_icassp18,chen+etal_slt18,holzenberger+etal_interspeech18,chung+glass_interspeech18,haque+etal_icassp19,shi+etal_arxiv19,palaskar+etal_icassp19,settle+etal_icassp19,jung+etal_asru19}.
For zero-resource settings, one approach is to train an unsupervised model on unlabelled data from the target language.
Chung et al.~\cite{chung+etal_interspeech16} trained an autoencoding encoder-decoder recurrent neural network (RNN) on unlabelled speech segments and used (a projection of) the final encoder hidden state as embedding.
Kamper~\cite{kamper_icassp19} extended this approach: instead of reconstructing an input segment directly, the correspondence autoencoder RNN (\system{CAE-RNN}) attempts to reconstruct another speech segment of the same type as the input.
Since labelled data isn't available for zero-resource languages, the input-output pairs for the \system{CAE-RNN} are obtained from an unsupervised term discovery (UTD) system, which automatically finds recurring word-like patterns in an unlabelled speech collection~\cite{park+glass_taslp08,jansen+vandurme_asru11}.

A recent alternative for obtaining embeddings on a zero-resource language is to use multilingual transfer learning~\cite{ma+etal_arxiv20,kamper+etal_icassp20,kamper+etal_arxiv2020,hu+etal_arxiv20}.
The idea is to train a supervised multilingual AWE model jointly on a number of well-resourced languages for which labelled data is available, but to then apply the model to an unseen zero-resource language.
This multilingual transfer approach was found to outperform monolingual unsupervised learning approaches in~\cite{kamper+etal_arxiv2020,hu+etal_arxiv20}.

One question is whether unsupervised learning and multilingual transfer are complementary.
More concretely, can multilingual transfer further benefit from incorporating unsupervised learning?
In this paper we answer this question by using unsupervised adaptation: a multilingual AWE model is updated by fine-tuning (a subset of) its parameters to a particular zero-resource language.
To obtain training targets, we use the same approach as for the unsupervised model in~\cite{kamper_icassp19}, and apply a UTD system to unlabelled data from the target language.
We consider unsupervised adaptation of multilingual \system{CAE-RNN} models, \system{SiameseRNN} models~\cite{settle+livescu_slt16}, and a new AWE approach based on self-supervised contrastive learning.

\textit{Self-supervised learning} involves using proxy tasks for which target labels can automatically be obtained from the data~\cite{doersch+zisserman_iccv17,asano+etal_iclr20}.
Originally proposed for vision problems~\cite{doersch+etal_iccv15,noroozi+favaro_eccv16,gidaris+etal_arxiv18}, it has since also been used as an effective 
pretraining step for supervised speech recognition~\cite{pascual+etal_arxiv19,synnaeve+etal_arxiv19,baevski+etal_iclr20,baevski+mohamed_icassp20,wang+etal_icassp20,ravanelli+etal_icassp20}.
It is somewhat difficult to distinguish self-supervised from unsupervised learning.\footnote{E.g., the unsupervised monolingual \tablesystem{CAE-RNN}~\cite{kamper_icassp19} is referred to as a self-supervised model in~\cite{algayres+etal_arxiv20}, since it fits the definition exactly: training targets are automatically obtained from the data for a reconstruction task.
}
But, importantly for us, a number of loss functions have been introduced in the context of self-supervised learning which have not been considered for AWEs.
Here we specifically consider the contrastive loss of~\cite{chen+etal_arxiv2020, sohn_nips2016}.
While a Siamese AWE model~\cite{kamper+etal_icassp16,settle+livescu_slt16} optimises the relative distance between one positive and one negative pair, our contrastive AWE model jointly embeds a number of speech segments and then
attempts to select a positive item from among several negative items.
We compare the \system{ContrastiveRNN} to \system{CAE-RNN} and \system{SiameseRNN} models in both the purely unsupervised monolingual and the supervised multilingual transfer settings.
We use an intrinsic word discrimination task on six languages (which we treat as zero-resource).

Our main contributions are as follows. (i)~For purely unsupervised monolingual AWEs, we show that a \system{ContrastiveRNN} using UTD segments as training targets outperforms previous unsupervised models by between 5\% and 19\% absolute in average precision (AP).
(ii)~We compare contrastive learning to other supervised AWE models for multilingual transfer (without adaptation)
and find that
the multilingual \system{ContrastiveRNN} only gives improvements on some
(but not all)
zero-resource languages compared to the multilingual \system{CAE-RNN} and \system{SiameseRNN}.
(iii)~However, when performing unsupervised adaptation, adapted multilingual \system{ContrastiveRNN}s outperform the other adapted models on five out of six zero-resource languages, with improvements of up to 12\% absolute in AP on some languages, resulting in the best reported results on these data sets.
(iv)~We perform probing experiments which show that the \system{ContrastiveRNN} is generally better at abstracting away from speaker identity.

\section{Acoustic word embedding models}
\label{sec:awe_models}

We first provide an overview of two existing acoustic word embedding
(AWE) models. 
We then introduce a new contrastive model.
Each of these models can be trained using labelled word segments (making them supervised) or by using discovered words from a UTD system (making them unsupervised); in this section we are agnostic to the training method, but we discuss how we use the different models in detail in Section~\ref{sec:embedding_methods}.

\subsection{Correspondence autoencoder RNN}
\label{ssec:cae}

\begin{figure}[!t]
	\centering
	{\includegraphics[scale=0.24]{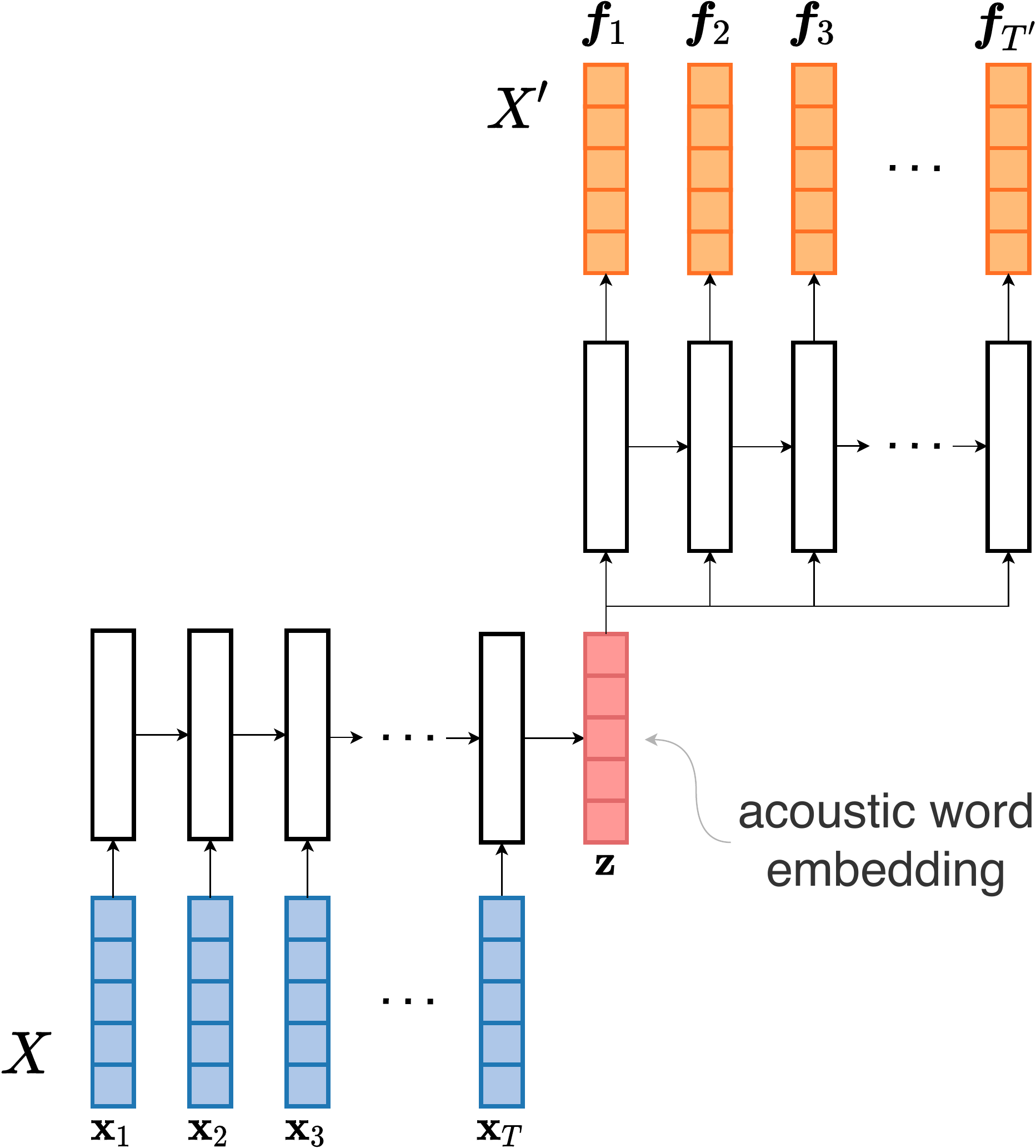}}
	\vspace*{-2pt}
	\caption{
	The \system{CAE-RNN} is trained to reconstruct an instance $X^\prime$ of the same word type as the 
	input sequence $X$.
	$T^\prime$ and $T$ are the lengths of $X^\prime$ and $X$, respectively. 
	}
	\label{fig:cae_rnn}
\end{figure}

The correspondence autoencoder recurrent neural network~(\system{CAE-RNN})~\cite{kamper_icassp19} is an extension of an autoencoder RNN~\cite{chung+etal_interspeech16}.
Both models consist of an encoder RNN and a decoder RNN.
The encoder produces a fixed-dimensional representation of a variable-length word segment which is then fed to the input of the decoder to reconstruct the original input sequence.
In the \system{CAE-RNN}, unlike the autoencoder, the target output is not identical to the input, but rather an instance of the same word type.
Figure~\ref{fig:cae_rnn} illustrates this model.
Formally, the \system{CAE-RNN} is trained on pairs of speech segments $(X, X^\prime)$, with $X = \mathbf{x}_1, \ldots, \mathbf{x}_T$ and $X^\prime = \mathbf{x}^{\prime}_1, \ldots, \mathbf{x}^{\prime}_T$, containing different instances of the same word type, with each $\mathbf{x}_t$ an acoustic feature vector.
The loss for a single training pair is therefore $J = \sum_{t=1}^{T^{\prime}} \norm{\mathbf{x}_t^{\prime} - \boldsymbol{f}_t(X)}^2$, where $\boldsymbol{f}_t(X)$ is the $t^{th}$ decoder output conditioned on the embedding $\mathbf{z}$.
The embedding $\mathbf{z}$ is a projection of the final encoder RNN hidden state. As in~\cite{kamper_icassp19}, we first pretrain the \system{CAE-RNN} as an autoencoder and then switch to the loss function for correspondence training.

\subsection{Siamese RNN}
\label{ssec:siamese}

\begin{figure}[!t]
	\centering
	{\includegraphics[scale=0.24]{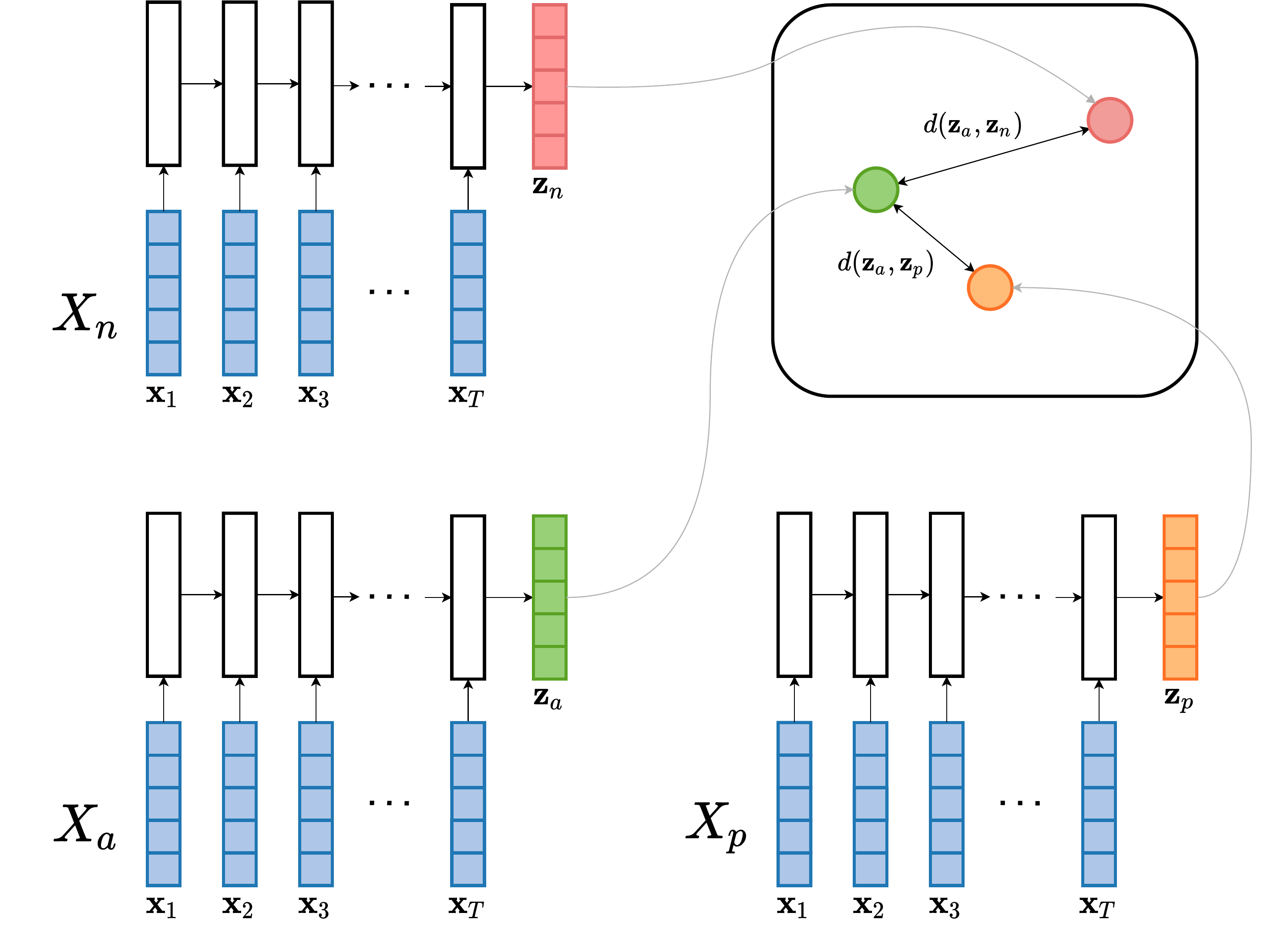}}
	\vspace*{-2pt}
	\caption{In the \system{SiameseRNN}, 
	three encoder RNNs 
	use the same set of parameters to produce embeddings $\mathbf{z}_a$, $\mathbf{z}_p$, $\mathbf{z}_n$ from input segments $X_a$, $X_p$, $X_n$. The model 
	is trained 
	to minimise the distance between the anchor and the positive item
	while 
	maximising
	the distance between the anchor 
	and 
	negative item.}
	\label{fig:siamese_rnn}
\end{figure}

Unlike the 
reconstruction loss used in 
the \system{CAE-RNN}, the \system{SiameseRNN} model explicitly optimises relative distances 
between embeddings~\cite{settle+livescu_slt16}.
Given input sequences $X_a$, $X_p$, $X_n$, the model produces embeddings $\mathbf{z}_a$, $\mathbf{z}_p$, $\mathbf{z}_n$, as illustrated in Figure~\ref{fig:siamese_rnn}.
Inputs $X_a$ and $X_p$ are from the same word type (subscripts indicate anchor and positive) and $X_n$ is from a different word type (negative).
For a single triplet of inputs, the model is trained using 
the triplet loss function,\footnote{Some studies~\cite{he+etal_iclr17,ng+lee_arxiv20,kamper+etal_arxiv2020} refer to this as a \textit{contrastive loss}, but we use \textit{triplet loss} here to explicitly distinguish it from the loss in Section~\ref{ssec:contrastive}.} defined as~\cite{weinberger+saul_jmlr09,chechik+etal_jmlr10}: $
J = \text{max} \{0, m + d(\mathbf{z}_a, \mathbf{z}_p) -  d(\mathbf{z}_a, \mathbf{z}_n)\}$, with
$m$ a margin parameter and $d(\mathbf{u}, \mathbf{v}) = 1 - \mathbf{u}^{\top}\mathbf{v}/\norm{\mathbf{u}}\norm{\mathbf{v}}$ denoting the cosine distance between two vectors $\mathbf{u}$ and $\mathbf{v}$. 
This loss is at a minimum when all embedding pairs $(\mathbf{z}_a , \mathbf{z}_p)$ of the same type are more similar by a margin $m$ than pairs $(\mathbf{z}_a , \mathbf{z}_n)$ of different types. 
To sample negative examples we use an online batch hard strategy~\cite{hermans+etal_arxiv2017}: for each item (anchor) in the batch we select 
the hardest positive and hardest negative example. 

\subsection{Contrastive RNN}
\label{ssec:contrastive}

\begin{figure}[!t]	
	\begin{minipage}[a]{0.45\linewidth}
		\centering
		\centerline{\includegraphics[width=0.95\linewidth]{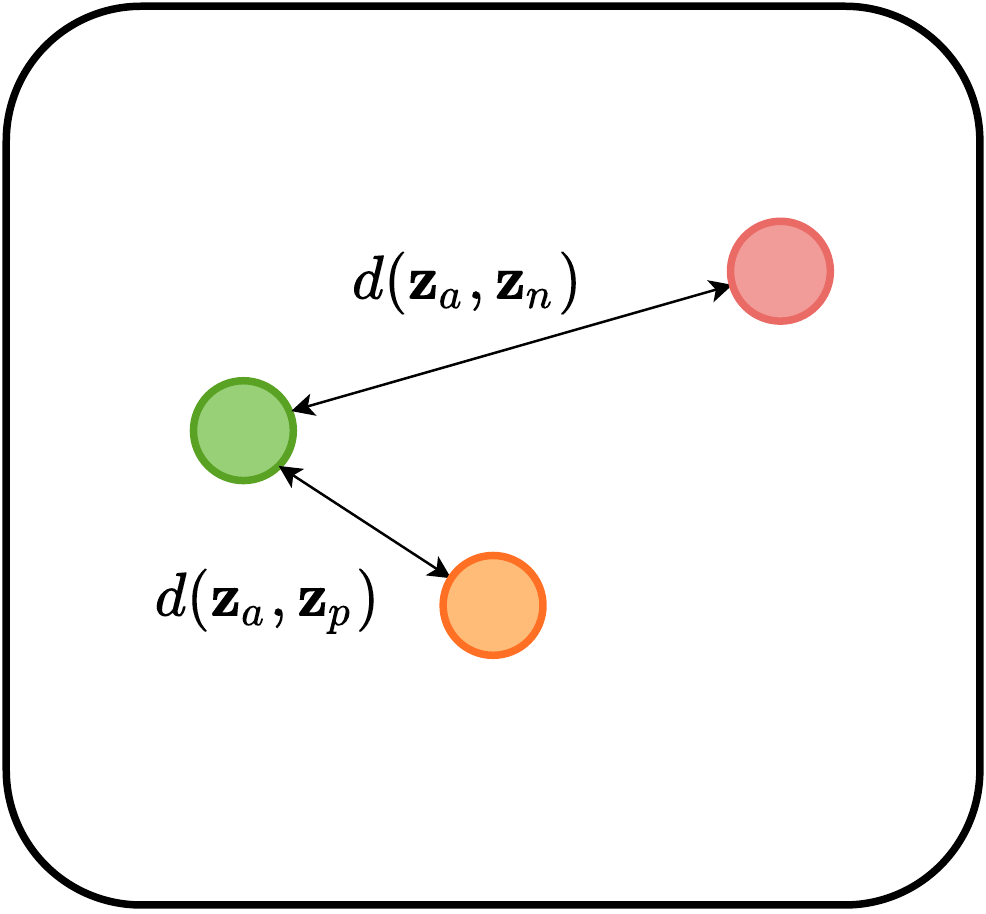}}
		\centerline{(a) Single negative example.}\medskip
	\end{minipage}
	\hfill
	\begin{minipage}[a]{0.45\linewidth}
		\centering
		\centerline{\includegraphics[width=0.95\linewidth]{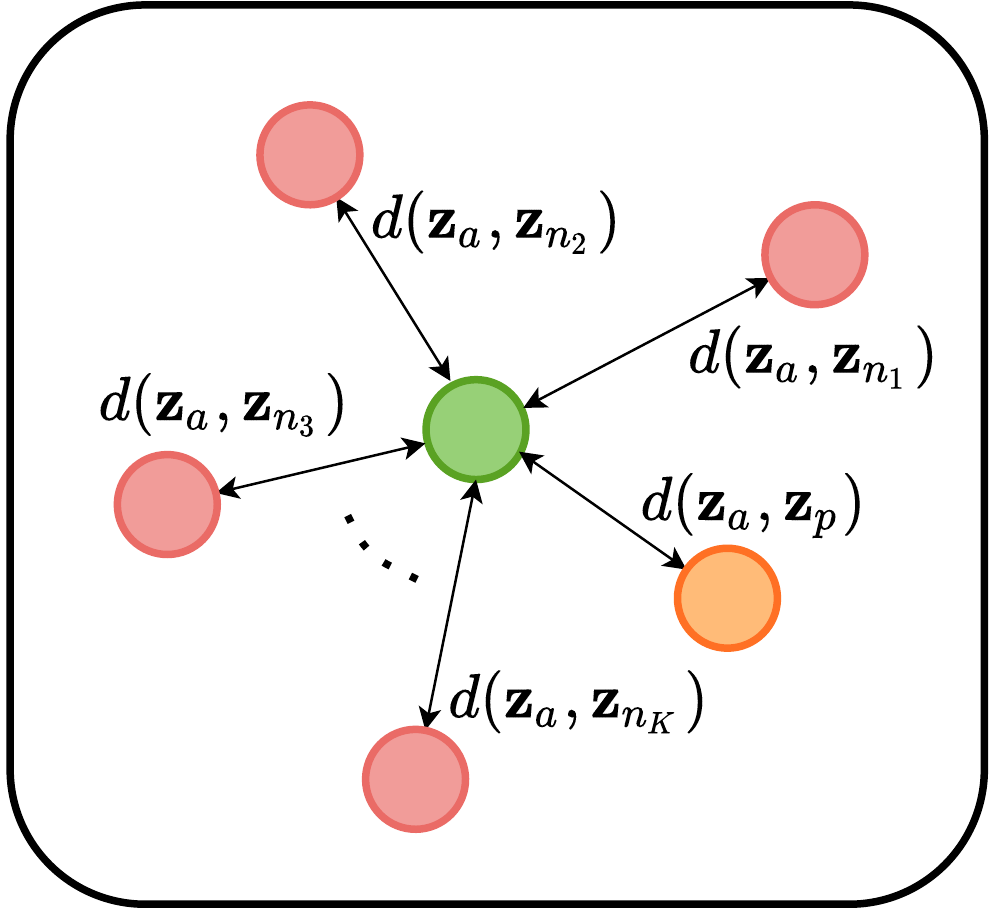}}
		\centerline{(b) Multiple negative examples.}\medskip
	\end{minipage}
	\vspace*{-2pt}
	\caption{A visualisation of the difference in the optimisation of (a) the \system{SiameseRNN} and (b) the \system{ContrastiveRNN} for a single positive pair $(\mathbf{z}_a, \mathbf{z}_p)$ in the embedding space.
	} 
	\label{fig:contrastive_rnn}
\end{figure}

As an extension of
the triplet loss function, 
we consider a loss
that incorporates multiple negative examples for each positive pair.
Concretely, given inputs $X_a$ and $X_p$ and multiple negative examples $X_{n_{1}}, \ldots, X_{n_{K}}$, the \system{ContrastiveRNN} produces embeddings $\mathbf{z}_a, \mathbf{z}_p, \mathbf{z}_{n_{1}}, \ldots, \mathbf{z}_{n_{K}}$. 
Let $\text{sim}(\mathbf{u}, \mathbf{v}) = \mathbf{u}^{\top}\mathbf{v}/\norm{\mathbf{u}}\norm{\mathbf{v}}$ denote the cosine similarity between two vectors $\mathbf{u}$ and $\mathbf{v}$.
The loss given a positive pair $(X_a, X_p)$ and the set of negative examples is then defined as~\cite{chen+etal_arxiv2020}:
\begin{equation*}
	J = -\text{log}\frac{\text{exp}\big\{\text{sim}(\mathbf{z}_a, \mathbf{z}_p)/\tau\big\}}{\sum_{j \in \{p, n_1, \hdots, n_K\}}^{}\text{exp}\big\{\text{sim}(\mathbf{z}_a, \mathbf{z}_j)/\tau\big\}}\,\text{,}
\label{eqn:contrastive_loss}
\end{equation*} 
where $\tau$ is a temperature parameter.
The difference between this loss and the triplet loss used in 
the \system{SiameseRNN} is illustrated in Figure~\ref{fig:contrastive_rnn}.
To sample negative examples 
we use an offline batch construction process. To construct 
a single batch, we choose 
$N$ distinct positive pairs. 
Given 
a positive pair $(X_a, X_p)$, the remaining $2(N-1)$ items 
are then treated as negative examples. 
The final loss is calculated as the sum of the loss over 
all $N$ positive pairs within the 
batch. 
As far as we are aware, the \system{ContrastiveRNN} has not been used as an AWE model in any previous work.

\section{Acoustic word embeddings for zero-resource languages}
\label{sec:embedding_methods}

In Section~\ref{sec:awe_models} we were agnostic to how training targets for the different AWE models are obtained.
In this section we describe different strategies for training AWE models, specifically for zero-resource languages where labelled data is not available.
One option is to train unsupervised monolingual models directly on unlabelled data (Section~\ref{ssec:monolingual}).
Another option is to train a supervised multilingual model on labelled data from well-resourced languages and then apply the model to a zero-resource language (Section~\ref{ssec:multilingual}).
All three of the AWE models in Section~\ref{sec:awe_models} can be used in both of these settings, as explained below.
Finally, in Section~\ref{ssec:multilingual_adapt} we describe a new combined approach where multilingual models are fine-tuned to a zero-resource language using unsupervised adaptation.

\subsection{Unsupervised monolingual models}
\label{ssec:monolingual} 
For any of the AWE models in Section~\ref{sec:awe_models}, we need pairs of segments containing words of the same type; for the \system{SiameseRNN} and \system{ContrastiveRNN} we additionally need negative examples.
In a 
zero-resource setting there
is no
transcribed speech to construct such pairs. 
But pairs can be
obtained automatically~\cite{jansen+etal_icassp13b}:
we apply an unsupervised term discovery (UTD) system~\cite{jansen+vandurme_asru11} to an
unlabelled speech collection from the target 
zero-resource language. 
This system discovers pairs of word-like 
segments, predicted to be of the same unknown type. 
The discovered 
pairs can be used to sample 
positive and negative examples for any of the three models in Section~\ref{sec:awe_models}.
Since the UTD system has no prior knowledge of the language or word boundaries within the unlabelled speech data, 
the entire process can be considered unsupervised.
Using this methodology, we consider purely unsupervised monolingual version of each of the three AWE models in Section~\ref{sec:awe_models}.

\subsection{Supervised multilingual models}
\label{ssec:multilingual}

Instead of relying on discovered words from the target zero-resource language, we can exploit labelled 
data from well-resourced languages to train a single multilingual supervised AWE model~\cite{kamper+etal_arxiv2020,hu+etal_arxiv20}.
This model can then be applied to an unseen zero-resource language.
Since a supervised model is trained for one task and applied to another, this can be seen as a form of \textit{transfer learning}~\cite{pan+yang_tkde09,ruder_phd19}.
We consider supervised multilingual variants of the three models in Section~\ref{sec:awe_models}.
Experiments in~\cite{kamper+etal_arxiv2020} showed that multilingual versions of the \system{CAE-RNN} and \system{SiameseRNN} outperform unsupervised monolingual variants.
A multilingual \system{ContrastiveRNN} hasn't been considered in a previous study, as far as we know.

\subsection{Unsupervised adaptation of multilingual models}
\label{ssec:multilingual_adapt}

While previous studies have found that multilingual AWE models (Section~\ref{ssec:multilingual}) are superior to unsupervised AWE models (Section~\ref{ssec:monolingual}), one question is whether multilingual models could be tailored to a particular zero-resource language in an unsupervised way.
We propose 
to adapt a 
multilingual AWE model to a target zero-resource language: a multilingual model's parameters (or a subset of the parameters) are fine-tuned
using 
discovered word pairs. 
These discovered segments are obtained by applying 
a UTD system 
to unlabelled data from the target zero-resource language.
The idea is that 
adapting the multilingual AWE model to the target language
would allow
the model to learn aspects
unique to
that
language.

We consider the adaptation of multilingual versions of all three AWE models in Section~\ref{sec:awe_models}.
On development data, we experimented with which parameters to update and which to keep fixed from the source multilingual model.
For the \system{CAE-RNN}, we found that it is best to freeze the multilingual encoder RNN weights
and only update the weights between the final encoder RNN hidden state and the embedding; we also found that it is best to re-initialise the decoder RNN weights randomly before training on the target language.
For the \system{SiameseRNN} and \system{ContrastiveRNN}, we update all weights during adaptation.

As far as we know, we are the first to perform \textit{unsupervised} adaptation of multilingual AWE models for the zero-resource setting.
However,~\cite{hu+etal_arxiv20} showed the benefit of \textit{supervised} adaptation, where (limited) labelled data from a target language is used to update the parameters of a multilingual AWE model.

\section{Experimental setup}
\label{sec:experiment_setup}

We perform 
experiments using 
the GlobalPhone corpus of read speech~\cite{schultz+etal_icassp13}.
As in~\cite{hermann+etal_csl20, kamper+etal_arxiv2020}, we
treat six languages as our target zero-resource languages: Spanish (ES), Hausa (HA), Croatian (HR), Swedish (SV), Turkish (TR) and Mandarin~(ZH).
Each language has on average 16 hours of training, 2 hours of development and 2 hours of test data.
We apply the UTD system of~\cite{jansen+vandurme_asru11} to the training set of each zero-resource language 
and use the discovered pairs to train unsupervised monolingual embedding models (Section~\ref{ssec:monolingual}). 
The UTD system discovers around 
36k pairs for each language,
where pair-wise matching precisions vary between $32\%$ (SV) and $79\%$ (ZH).
Training conditions for the unsupervised monolingual \system{CAE-RNN}, \system{SiameseRNN} and \system{ContrastiveRNN} models are determined by doing validation on the Spanish
development data. 
The same hyperparameters 
are then used 
for the five remaining zero-resource languages.

For training supervised multilingual embedding models (Section~\ref{ssec:multilingual}), six other GlobalPhone languages are chosen as well-resourced languages: Czech, French, Polish, Portuguese, Russian and Thai.
Each well-resourced language has on average 21 hours of labelled training data. 
We pool the data from all six well-resourced languages and train a multilingual \system{CAE-RNN}, a \system{SiameseRNN} and a \system{ContrastiveRNN}.
Instead of using the development data from one of the zero-resource languages, we 
use another well-resourced language, German, 
for validation of each 
model before applying it to the zero-resource languages. 
We only use 300k positive word pairs for each model, as further increasing the number of pairs did not 
give improvements on the German validation data.

As explained in Section~\ref{ssec:multilingual_adapt}, we adapt each of the multilingual models to each of the six zero-resource languages using the same discovered pairs as for the unsupervised monolingual models.
We again use Spanish development data to determine hyperparameters. 


All speech audio is parametrised as $D = 13$ dimensional static Mel-frequency cepstral coefficients (MFCCs).
All our models have 
a similar architecture: encoders and decoders consist of three unidirectional RNNs with 400-dimensional hidden vectors, and all models use an embedding size of 130 dimensions.
Models are optimised using Adam optimisation~\cite{kingma+ba_iclr15}. 
The margin parameter $m$ in Section \ref{ssec:siamese} and  temperature parameter $\tau$ in Section \ref{ssec:contrastive} are 
set to $0.25$ and $0.1$, respectively. 
We implement all our models 
in PyTorch.\footnote{ \url{https://github.com/christiaanjacobs/globalphone_awe_pytorch}}

We use a word discrimination task~\cite{carlin+etal_icassp11} to measure the intrinsic quality of the resulting AWEs.
To evaluate a particular AWE model, a set of isolated test word segments is embedded.
For every word pair in this set, the cosine distance between their embeddings is calculated.
Two words can then be classified as being of the same or different type based on some distance threshold, and a precision-recall curve is obtained by varying the threshold.
The area under this curve is used as final evaluation metric, referred to as the average precision~(AP).
We are particularly interested in obtaining embeddings that are speaker invariant. 
We therefore calculate AP by only taking the recall over instances of the same word spoken by different speakers, i.e.\ we consider the more difficult setting where a model does not get credit for recalling the same word if it is said by the same speaker.
\section{Experimental results}
\label{sec:results}

We start in Section~\ref{ssec: word_discrimination} by evaluating the different AWE models using the intrinsic word discrimination task described above.
Instead of only looking at word discrimination results, it is useful to also use other methods to try and better understand the organisation of AWE spaces~\cite{matusevych+etal_baics20}, especially in light of recent results~\cite{algayres+etal_arxiv20} showing that AP has limitations.
We therefore look at speaker classification performance in Section~\ref{ssec: speaker_invariance}, and give a qualitative analysis of adaptation in Section~\ref{ssec: qualitative_analysis}.


\subsection{Word discrimination}
\label{ssec: word_discrimination}
We first consider purely unsupervised monolingual models (Section~\ref{ssec:monolingual}).
We are particularly interested in the performance of the \system{ContrastiveRNN}, which has not been considered in previous work.
The top section in Table~\ref{tbl:multi} shows the 
performance for 
the unsupervised monolingual AWE models applied to the test data from the six zero-resource languages.\footnote{We note that the results for the \tablesystem{CAE-RNN} and \tablesystem{SiameseRNN} here are slightly different to that of~\cite{kamper+etal_icassp20,kamper+etal_arxiv2020}, despite using the same test and training setup. We believe this is due to the different negative sampling scheme for the \tablesystem{SiameseRNN} and other small differences in our implementation.}
As a baseline, we also give the results where DTW is used directly on the MFCCs to perform the word discrimination task.
We see that the \system{ContrastiveRNN} consistently outperforms the \system{CAE-RNN} and \system{SiameseRNN}
approaches on all six zero-resource languages.
The \system{ContrastiveRNN} is also the only model to perform better than DTW on all six zero-resource languages, which is noteworthy since DTW has access to the full sequences for discriminating between words.

\begin{table}[!b]
    \mytable
    \caption{AP (\%) on test data for the six zero-resource languages.
    The purely unsupervised monolingual models are trained on discovered word segments (top).
    The multilingual models are trained on true words by pooling labelled training data from six 
    well-resourced languages (middle). These 
    models are adapted to each of the languages using discovered words from the target zero-resource language (bottom).
	The best approach in each subsection is shown in bold.}
    \vspace{5pt}
	\small
    \begin{tabularx}{1\linewidth}{@{}L *6{S[table-format=2.1]}@{}}
        \toprule
        Model & {ES} & HA & HR & SV & TR & ZH \\
        \midrule
        \underline{\textit{Unsupervised models:}} \\[2pt]        
	    \tablesystem{DTW} &36.2 &23.8 &17.0 &27.8 &16.2 &35.9 \\
		\tablesystem{CAE-RNN} &52.7 &18.6 &24.5 &28.0 &14.2 &33.7 \\
		\tablesystem{SiameseRNN} &56.6 &16.8 &21.1 &31.8 &22.8 &52.0 \\
		\tablesystem{ContrastiveRNN} &\textbf{70.6} &\textbf{36.4} &\textbf{27.8} &\textbf{37.9} &\textbf{31.3} &\textbf{57.1} \\[2pt]
        \underline{\textit{Multilingual models:}} \\[2pt]
        \tablesystem{CAE-RNN} &72.4 &49.3 &44.5 &\textbf{52.7} &34.4 &\textbf{53.9} \\
        \tablesystem{SiameseRNN} &70.3 &45.3 &40.6 &47.5 &27.7 &49.9 \\
        \tablesystem{ContrastiveRNN} &\textbf{73.3} &\textbf{50.6} &\textbf{45.1} &46.4 &\textbf{34.6} &53.2 \\[2pt]
        \underline{\textit{Multilingual adapted:} 
        } \\[2pt]
        \tablesystem{CAE-RNN} &74.2 &49.4 &\textbf{45.9} &53.4 &34.5 &53.9 \\
        \tablesystem{SiameseRNN} &74.5 &44.7 &37.6 &50.3 &30.3 &57.3 \\
        \tablesystem{ContrastiveRNN} &\textbf{76.6} &\textbf{56.7} &44.4 &\textbf{54.4} &\textbf{40.5} &\textbf{60.4} \\
        \bottomrule
    \end{tabularx}
    \label{tbl:multi}
\end{table}

Next, we consider the supervised multilingual models (Section~\ref{ssec:multilingual}). 
The middle section of Table~\ref{tbl:multi} shows the performance for the supervised multilingual models 
applied to the 
six zero-resource languages. 
By comparing these supervised multilingual models to the unsupervised monolingual models (top), 
we see that in almost all cases the multilingual models outperform the purely unsupervised monolingual models, as also in~\cite{kamper+etal_icassp20,kamper+etal_arxiv2020}.
However, 
on Mandarin, the unsupervised monolingual \system{ContrastiveRNN} model outperforms all three multilingual models.
Comparing the three multilingual models, we do not see a consistent winner between the \system{ContrastiveRNN} and 
\system{CAE-RNN}, with one performing better on 
some languages while the other 
performs better on others. 
The multilingual \system{SiameseRNN} generally performs worst, although it outperforms the \system{ContrastiveRNN} on Swedish.

Finally, we consider adapting the supervised multilingual models  (Section~\ref{ssec:multilingual_adapt}).
The results after adapting each multilingual model to each of the 
zero-resource languages are shown in the bottom section of
Table~\ref{tbl:multi}.
Comparing the middle and bottom sections of the table, we see that most of the adapted models outperform their corresponding source multilingual models, with the \system{ContrastiveRNN} and \system{SiameseRNN} improving substantially after adaptation on some of the languages.
The adapted \system{ContrastiveRNN} models outperform the adapted \system{CAE-RNN} and \system{SiameseRNN} models on five out of the six zero-resource languages, achieving some of the best reported results on these data sets~\cite{ann+etal_csl20,kamper+etal_icassp20}.
We conclude that unsupervised adaptation of multilingual models to a target zero-resource language is an effective AWE approach, especially when coupled with the self-supervised contrastive loss.

\begin{table}[!b]
    \mytable
    \caption{AP (\%) on development data.
    The supervised monolingual models are trained on ground truth words from the target languages to determine an upper bound on performance.
    The adapted models are the same as those in Table~\ref{tbl:multi} but applied to development data here for the purpose of analysis.
	}
    \vspace{5pt}
	\small
    \begin{tabularx}{1\linewidth}{@{}L *6{S[table-format=2.1]}@{}}
        \toprule
        Model & {ES} & HA & HR & SV & TR & ZH \\
        \midrule
        \underline{\textit{Supervised monolingual:}} \\[2pt]
		\tablesystem{CAE-RNN} &70.2 &79.7 &63.0 &55.8 &65.6 &84.2  \\
		\tablesystem{SiameseRNN} &78.6 &85.2 &79.3 &68.6 &77.6 &93.1 \\
		\tablesystem{ContrastiveRNN} &81.8 &82.4 &80.3 &70.9 &80.5 &92.1 \\[2pt]
        \underline{\textit{Multilingual adapted:}} \\[2pt]
        \tablesystem{CAE-RNN} &51.7 &59.5 &44.3 &38.6 &40.9 &52.2 \\
        \tablesystem{SiameseRNN} &51.5 &52.7 &38.4 &40.3 &33.0 &56.5 \\
        \tablesystem{ContrastiveRNN} &58.0 &60.5 &40.5 &43.8 &46.9 &60.2 \\
        \bottomrule
    \end{tabularx}
    \label{tbl:top-line}
\end{table}

One question is whether adapted models close the gap between the zero-resource setting and the best-case scenario where we have labelled data available in a target language.
To answer this, Table~\ref{tbl:top-line} compares multilingual models (bottom) to ``oracle'' supervised monolingual models trained on labelled data from the six evaluation languages (top) on development data.
Although Table~\ref{tbl:multi} shows that adaptation greatly improves performance in the zero-resource setting, Table~\ref{tbl:top-line} shows that multilingual adaptation still does not reach the performance of supervised monolingual models.


\subsection{Speaker classification}
\label{ssec: speaker_invariance}

\begin{table}[!t]
    \mytable
    \caption{
    Speaker classification accuracy (\%) on development data for the zero-resource languages using purely unsupervised monolingual models (top), multilingual models before adaptation (middle), and models after adaptation (bottom).
	}	
    \vspace{5pt}
	\small
    \begin{tabularx}{1\linewidth}{@{}L *6{S[table-format=2.1]}@{}}
        \toprule
        Model & {ES} & HA & HR & SV & TR & ZH \\
        \midrule
        \underline{\textit{Unsupervised models:}} \\[2pt]        
        \tablesystem{CAE-RNN} & 52.8 & 49.5 & 53.8 & 47.8 & 48.9 & 62.0 \\
        \tablesystem{SiameseRNN} & 38.9 & 38.4 & \textbf{37.2} & 36.4 & 38.6 & 44.3 \\
        \tablesystem{ContrastiveRNN} & \textbf{34.6} & \textbf{33.1} & 37.8 & \textbf{33.6} & \textbf{35.5} & \textbf{39.9} \\[2pt]
        \underline{\textit{Multilingual models:}} \\[2pt]        
        \tablesystem{CAE-RNN} & 40.6 & 44.7 & 45.0 & 42.3 & 42.0 & 49.4 \\
        \tablesystem{SiameseRNN} & 32.6 & 30.3 & 31.9 & \textbf{28.6} & \textbf{30.3} & 32.5 \\ 
        \tablesystem{ContrastiveRNN} & \textbf{27.6} & \textbf{28.7} & \textbf{26.9} & 29.2 & 32.1 & \textbf{31.9} \\[2pt]
        \underline{\textit{Multilingual adapted:}} \\[2pt]
        \tablesystem{CAE-RNN} & 40.3 & 45.3 & 45.0 & 43.1 & 41.0 & 49.7 \\
        \tablesystem{SiameseRNN} & 33.6 & 33.9 & 30.6 & \textbf{29.4} & \textbf{33.7} & 37.9 \\
        \tablesystem{ContrastiveRNN} & \textbf{30.3} & \textbf{32.1} & \textbf{26.9} & 29.5 & 36.5 & \textbf{35.1} \\        
        \bottomrule
    \end{tabularx}
    \label{tbl:monolingual_speaker}
\end{table}


To what extent do the different AWEs capture speaker information?
How does adaptation affect speaker invariance?
To measure speaker invariance, 
we use a linear classifier to predict a word's speaker identity from its AWE.
Specifically, we train a multi-class logistic regression model on 80\% of the development data and test it on the remaining 20\%.

The top section of Table~\ref{tbl:monolingual_speaker} shows speaker classification results on development data for the three types of monolingual unsupervised models (Section~\ref{ssec:monolingual}).
Since we are interested in how well models abstract away from speaker information, we consider lower accuracy as better (shown in bold).
The \system{ContrastiveRNN} achieves the lowest speaker classification performance across all languages, except on Croatian where it performs very similarly to the \system{SiameseRNN}.
This suggests that among the unsupervised monolingual models, the \system{ContrastiveRNN} is the best at abstracting away from speaker identity (at the surface level captured by a linear classifier).

Next, we consider speaker classification performance for the multilingual models (Section~\ref{ssec:multilingual}).
Comparing the middle and top sections of Table~\ref{tbl:monolingual_speaker}, we see that for each multilingual model (middle) the speaker classification performance drops from its corresponding unsupervised monolingual version (top) across all six languages, again indicating an improvement in speaker invariance. 
Comparing the three multilingual models to each other (middle), the \system{ContrastiveRNN} has the lowest speaker classification performance on four out of the six evaluation languages. 

Finally, we look at the impact of unsupervised adaptation (Section~\ref{ssec:multilingual_adapt}) on speaker invariance, shown at the bottom of Table~\ref{tbl:monolingual_speaker}. 
After adaptation (bottom) we see that speaker classification results improve consistently compared to their corresponding source multilingual model (middle).
Although this seems to indicate that the adapted AWEs capture more speaker information, these embeddings still lead to better word discrimination performance (Table~\ref{tbl:multi}).
A similar trend was observed in~\cite{kamper+etal_arxiv2020}: a model leading to better (linear) speaker classification performance does not necessarily give worse AP.
Recent results in~\cite{algayres+etal_arxiv20} showed that AP is limited in its ability to indicate downstream performance for all tasks. 
Further analysis is required to investigate this seeming contradiction.

Importantly for us, it seems that unsupervised adaptation using unlabelled data in a target zero-resource language leads to representations which better distinguish between speakers in that language. 

\subsection{Qualitative analysis of adaptation}
\label{ssec: qualitative_analysis}

\begin{figure}[!t]	
	\begin{minipage}[a]{.49\linewidth}
		\centering
		\centerline{\includegraphics[width=\linewidth]{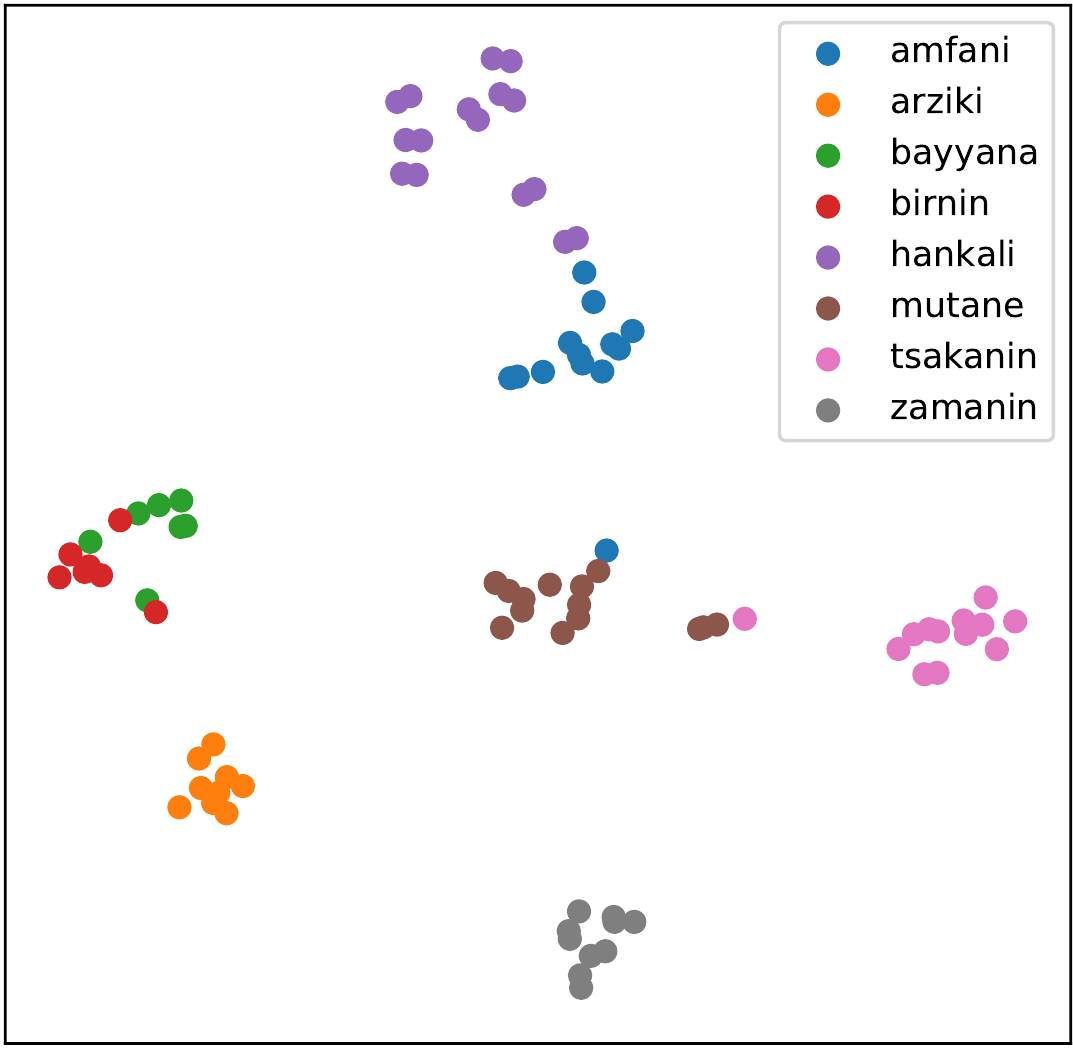}}
		\centerline{(a) Before adaptation.}\medskip
	\end{minipage}
	\hfill
	\begin{minipage}[a]{0.49\linewidth}
		\centering
		\centerline{\includegraphics[width=\linewidth]{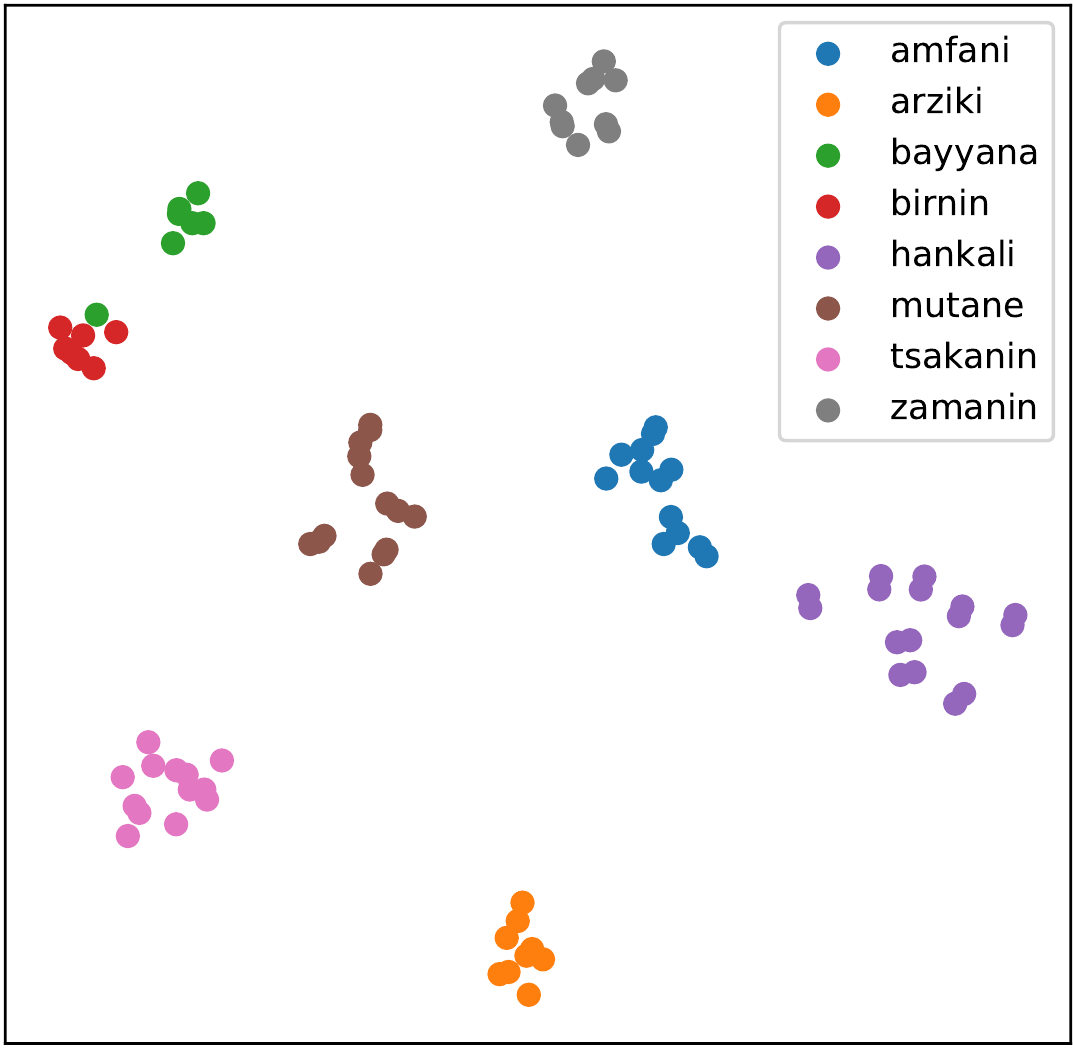}}
		\centerline{(b) After adaptation.}\medskip
	\end{minipage}
	\vspace*{-2pt}
	\caption{t-SNE visualisations 
		of acoustic embeddings for 
		the most frequent 
		words 
		in the Hausa data, 
		produced by (a) the multilingual \system{ContrastiveRNN} model and (b) the multilingual \system{ContrastiveRNN} model adapted to Hausa.} 
	\label{fig:tsne}
\end{figure}

Figure~\ref{fig:tsne} shows 
t-SNE visualisations~\cite{maaten+hinton_jmlr2008} of the AWEs produced by 
the
\system{ContrastiveRNN} on Hausa data
before and after adaptation. 
In this curated example, we see how some of the words that are clustered together by the multilingual model (e.g. ``amfani'' and ``hankali'') are separated after adaptation.

\section{Conclusion}
\label{sec:conclusion}

We have compared a self-supervised contrastive acoustic word embedding approach to two existing methods in a word discrimination task on six zero-resource languages.
In a purely unsupervised setting where
words from a term discovery system are used for self-supervision, the contrastive model outperformed unsupervised correspondence autoencoder and Siamese embedding models.
In a multilingual transfer
setting 
where a model is trained on several well-resourced languages and then applied to a zero-resource language, the contrastive model didn't show consistent improvements.
However, it performed best in a setting where multilingual models are adapted to a particular zero-resource language using the unsupervised discovered word segments, leading to the best reported results on this data.
Analysis shows that the contrastive approach abstracts away from speaker identity more than the other two approaches.
Future work will involve extending our analysis and performing comparative experiments in a downstream query-by-example search task.

\newpage
\bibliography{mybib}

\end{document}